\documentclass{llncs}

\usepackage{array}
\usepackage{graphicx}
\usepackage{algorithm}
\usepackage[noend]{algorithmic}

\title{Scalable Dictionary Classifiers for Time Series Classification}
\author{Matthew Middlehurst, William Vickers and Anthony Bagnall}
\date{December 2018}

\institute{
School of Computing Sciences, University of East Anglia, UK\\
\email{M.Middlehurst@uea.ac.uk}
}

\begin{document}

\maketitle

\begin{abstract}
Dictionary based classifiers are a family of algorithms for time series classification (TSC), that focus on capturing the frequency of pattern occurrences in a time series.
The ensemble based Bag of Symbolic Fourier Approximation Symbols (BOSS) was found to be a top performing TSC algorithm in a recent evaluation, as well as the best performing dictionary based classifier.
A recent addition to the category, the Word Extraction for Time Series Classification (WEASEL), claims an improvement on this performance.
Both of these algorithms however have non-trivial scalability issues, taking a considerable amount of build time and space on larger datasets.
We evaluate changes to the way BOSS chooses classifiers for its ensemble, replacing its parameter search with random selection.
This change allows for the easy implementation of contracting, setting a build time limit for the classifier and check-pointing, saving progress during the classifiers build.
To differentiate between the two BOSS ensemble methods we refer to our randomised version as RBOSS.
Additionally we test the application of common ensembling techniques to help retain accuracy from the loss of the BOSS parameter search.
We achieve a significant reduction in build time without a significant change in accuracy on average when compared to BOSS by creating a size $n$ weighted ensemble selecting the best performers from $k$ randomly chosen parameter sets.
Our experiments are conducted on datasets from the recently expanded UCR time series archive.
We demonstrate the usability improvements to RBOSS with a case study using a large whale acoustics dataset for which BOSS proved infeasible. 


\end{abstract} 

\section{Introduction}

Dictionary based learning is commonly employed in signal processing, computer vision and audio processing to capture recurring discriminatory features. The approach has been successfully applied to time series classification (TSC) in a variety of ways. An extensive experimental study~\cite{bagnall17bakeoff} found that the best dictionary approach on the University of California, Riverside datasets~\cite{UCRWeb} was the ensemble classifier the Bag of Symbolic Fourier Approximation Symbols (BOSS). It was shown that dictionary based classifiers detect a fundamentally different type of discriminatory features than other TSC approaches~\cite{bagnall17bakeoff}, and the addition of the BOSS ensemble to the meta ensemble the Hierarchical Vote Collective of Transformation-based Ensembles (HIVE-COTE)~\cite{lines18hive} leads to a significant improvement in accuracy. The BOSS transform works by running a sliding window of fixed length along each series, then shortening and discretising each window to form a combination of discrete symbols (referred to as a word). It counts the frequency of each unique word, to form a histogram of word counts. It has three parameters: the word length $l$, the alphabet size $\alpha$ and the window length $w$. A single BOSS base classifier stores the histogram for a fixed set of parameter values, then classifies new instances with a nearest neighbour classifier. The BOSS ensemble evaluates a range of parameter combinations over a grid, then retains all classifiers that are  within 92\% of the best combination, as measured by a leave-one-out cross-validation on the train data. Although it is an effective technique, the BOSS ensemble has some drawbacks. Firstly, the need to cross-validate each parameter combination means that it scales poorly. Secondly, the fact that it retains a variable number of base classifiers means that it is often very memory intensive. Thirdly, the histograms can be very large, so storing them all for each base classifier also requires a significant memory commitment for large problems. One proposed method to solve this, the BOSS vector space (BOSS-VS) classifier~\cite{schafer16boss-vs}, has been shown to be significantly less accurate than the full BOSS~\cite{schafer2017fast,lucas2018proximity}. We investigate whether we can mitigate against these problems without this significant loss in accuracy.  Patrick Sch{\"a}fer, the author of the BOSS paper, recently proposed an alternative dictionary based classifier called WEASEL~\cite{schafer2017fast}. WEASEL is claimed to be a more accurate approach than BOSS, not more efficient. It was proposed that WEASEL could be a possible replacement for BOSS in HIVE-COTE. We assess whether this is worthwhile and compare WEASEL to the BOSS variants we consider. Our primary contribution is to show that we can randomise the BOSS ensemble parameter search, thus avoid the need to grid search the parameter space, without significant loss of accuracy on average. We also show the merit of taking a subsample of the train data for each ensemble member, substantially lowering the classifiers time and space complexity.
Our further motivation for the enhancements to the BOSS ensemble is to improve the usability of HIVE-COTE. To this end, we aim to make each component of HIVE-COTE, including BOSS, both contractable (i.e. be able to build the best possible classifier in a fixed amount of time) and check-pointable (i.e. the build classifier stage can be stopped and restarted). We have released a Weka compatable version of BOSS that is integrated into the UEA Codebase\footnote{https://github.com/TonyBagnall/uea-tsc}. There is also have a python version of BOSS in the Alan Turing Institute sktime package\footnote{https://github.com/alan-turing-institute/sktime}.

The UCR TSC repository has recently been expanded~\cite{dau18archive} from 85 to 128 datasets available and including more complex problems such as series of differing length and series with missing values.
We make use of this expanded repository in our experiments but exclude datasets with missing values and variable length series due to the currently inability of included classifiers to handle such data.
The rest of this paper is structured as follows. Section~\ref{background} gives a detailed description of dictionary based classification algorithms BOSS and WEASEL. Section~\ref{enhancements} describes the alterations we make to the BOSS algorithm for our experiments, as well as the best performer from our results. 
Section~\ref{results} presents the results of of our experimental evaluation, and Section~\ref{conc} concludes and offers some ideas for future work.

\section{Time Series Dictionary Based Classifiers}
\label{background}
Classifiers that use frequency of words as the basis for finding discriminatory features are often referred to as dictionary based classifiers~\cite{bagnall17bakeoff}. They have close similarities to bag-of-words based approaches that are commonly used in computer vision. Informally, a dictionary based approach will be useful when the discriminatory features are repeating patterns that occur more frequently in one class then other classes. Generally, a base classifier takes a window of a real valued series, truncates it to form a shorter real valued series then discretises the shorter series to form a word. Histograms of words are formed from all windows. An in depth ablative study into the importance of these components can be found in~\cite{large19dictionary}.

    \subsection{Bag of SFA symbols (BOSS)}
    
A single BOSS~\cite{schafer15boss} base classifier proceeds as follows. For each series, it extracts the windows sequentially, normalising the window if the parameter $p$ is true. It then applies a Discrete Fourier Transform (DFT) to the resulting subseries, ignoring the first coefficient if $p$ is true. The DFT coefficients are truncated to include only the first $l/2$ Fourier terms (both real and imaginary). The truncated samples are then discretised into $\alpha$ possible values using an algorithm called Multiple Coefficient Binning (MCB) (see~\cite{schafer15boss}). MCB involves a preprocessing step to find the discretising break points by estimating the distribution of the Fourier coefficients. Consecutive windows producing the same word are only counted as a single instance of the word. This removes the confounding influence of long periods of little change in a series. A bespoke BOSS distance function is used with a nearest neighbour classifier to classify new instances. The distance function is non-symmetrical, only including the distance for features that are non-zero in the first feature vector given. A base BOSS classifier is described more formally in Algorithm~\ref{boss}.

        \begin{algorithm}[]
        	\caption{buildBaseBOSS(A list of $n$ time series of length $m$, ${\bf T}=({\bf X,y})$)}
        	\label{boss}
        	\begin{algorithmic}[1]
        \REQUIRE the word length $l$, the alphabet size $\alpha$, the window length $w$, normalisation parameter $p$
        		\STATE Let ${\bf H}$ be a list of $n$ histograms $({\bf h}_1,\ldots,{\bf h}_n)$
        		\STATE Let ${\bf B}$ be a matrix of $l$ by $\alpha$ breakpoints found by MCB
        		\FOR {$i \leftarrow  1$ to $n$}
        			\FOR {$j \leftarrow 1$ to $m-w+1$}
        				\STATE ${\bf s}\leftarrow x_{i,j} \ldots x_{i,j+w-1}$
        				\IF{$p$}
        				    \STATE $s \leftarrow $normalise($s$)
        				\ENDIF
        				\STATE ${\bf q} \leftarrow$ DFT(${\bf s}, l, \alpha$,$p$) \COMMENT{ {\em {\bf q} is a vector of the complex DFT coefficients}}
        				\IF{$p$}
            			    \STATE ${\bf q'} \leftarrow (q_2 \ldots q_{l/2+1})$
            			\ELSE
            			    \STATE ${\bf q'} \leftarrow (q_1 \ldots q_{l/2})$
            			\ENDIF
        				\STATE ${\bf r} \leftarrow$ SFAlookup(${\bf q', B}$)
        				\IF{${\bf r} \neq {\bf p}$}
        					\STATE $pos \leftarrow $index(${\bf r}$)
        					\STATE ${h}_{i,pos} \leftarrow {h}_{i,pos} + 1$
        				\ENDIF
        				\STATE ${\bf p} \leftarrow {\bf r} $
        			\ENDFOR
        		\ENDFOR
        	\end{algorithmic}
        \end{algorithm}

The BOSS base classifier has four parameters: window length $w$,  word length $l$, whether to normalise each window $p$ and alphabet size $\alpha$. The BOSS ensemble (also referred to as just BOSS), evaluates all BOSS base classifiers in the parameter space shown in table~\ref{BOSSparameters}.
This parameter search is used to determine which base classifiers are used in the ensemble.
Following~\cite{schafer15boss}, the alphabet size is fixed to 4 for all experiments. The number of window sizes is a function of the series length $m$. All BOSS base classifiers with a training accuracy within 92\% of the best performing base classifier are kept for the ensemble.
 This dependency on series length and variability of ensemble size is a factor that can significantly impact on efficiency.     Classification of new instances is then done using majority vote from the ensemble.

        \begin{table}[htb]
            \centering
			\caption{Range of values used in the BOSS parameter search. $m$ represents the length of the time series.}
			\begin{tabular}{c c c}
				Parameter & Num Values & Range \\
				\hline
				$w$ & $m$/4 & Linearly spaced from 10 to $m$  \\
				$l$ & 5 & \{16, 14, 12, 10, 8\} \\
				$mean$ & 2 & \{true, false\} \\
				$\alpha$ & 1 & \{4\} \\
			\end{tabular}
			\label{BOSSparameters}
		\end{table}

    \subsection{Word Extraction for Time Series Classification (WEASEL)}

    WEASEL~\cite{schafer2017fast} is a dictionary based classifier that is an extension of BOSS. WEASEL is a single classifier rather than an ensemble. WEASEL concatenates histograms for a range of parameter values of $w$ and $l$, then performs a feature selection to reduce the feature space. 
    
    Like BOSS, WEASEL performs a Fourier transform on each window.  DFT coefficients are no longer truncated and instead the most discriminative real and imaginary features are retained, as determined by an ANOVA F-test. The retained values are then discretised into words using information gain binning, similar to the MCB step in BOSS.
    WEASEL does not remove adjacent duplicate words as BOSS does (line 14 in Algorithm~\ref{boss}).
   The word and window size are used as keys to index the histogram. A further histogram is formed for bi-grams. An example of the resulting histogram bins would be `50 ac' for a unigram of window length 50 or `75 ac aa' for a bigram of window length 75. This generates a very large feature space. 
    The number of features is reduced using a chi-squared test after the histograms for each instance are created, removing any words which score below a threshold.
    The nearest neighbour classification and BOSS distance measure have been replaced with a logistic regression classifier to make the predictions for new cases.
    
    One of the main goals of WEASEL is to improve prediction time compared to other top performing classifiers while retaining accuracy. Hence, more resources are spent in training to find the most discriminative features for use in a single classifier. While prediction time is not covered in our results, it should be noted that in its results WEASEL is superior to BOSS and other ensemble based techniques in this respect~\cite{schafer2017fast}.
    WEASEL performs a parameter search for $mean$ and a reduced range of $l$ and uses a 10-fold cross-validation to determine the performance of each set.
    The alphabet size  $\alpha$ is fixed to 4 and new $chi$ parameter is fixed to 2.
    The building process for WEASEL is described in Algorithm~\ref{weasel}. 
    
        \begin{algorithm}[]
    	\caption{buildClassifierWEASEL(A list of $n$ cases of length $m$, ${\bf T}=({\bf X,y})$)}
    	\label{weasel}
    	\begin{algorithmic}[1]
    	\REQUIRE the word length $l$, the alphabet size $\alpha$, the maximal window length $w_{max}$, mean normalisation parameter $p$
    		\STATE Let ${\bf H}$ be the histogram ${\bf h}$
    		\STATE Let ${\bf B}$ be a matrix of $l$ by $\alpha$ breakpoints found by MCB using information gain binning
    		\FOR {$i \leftarrow  1$ to $n$}
    			\FOR {$w \leftarrow  2$ to $w_{max}$}
    				\FOR {$j \leftarrow 1$ to $m-w+1$}
    					\STATE ${\bf o}\leftarrow x_{i,j} \ldots x_{i,j+w-1}$
    					\STATE ${\bf q} \leftarrow$ DFT($o, w, p$) \COMMENT{ {\em {\bf q} is a vector of 	the complex DFT coefficients}}
    					\STATE ${\bf q'} \leftarrow$ ANOVA-F($q, l, y$) \COMMENT{ {\em use only the {\bf l} most discriminative ones}}					
    					\STATE ${\bf r} \leftarrow$ SFAlookup(${\bf q', B}$)  
    					\STATE $pos \leftarrow $index(${\bf w, r}$)
    					\STATE ${h}_{i,pos} \leftarrow {h}_{i,pos} + 1$
    				\ENDFOR					
    			\ENDFOR
    		\ENDFOR		
    		\STATE $h \leftarrow \chi^2(h, y)$ \COMMENT{ {\em feature selection using the chi-squared test} }
    		\STATE fitLibLinear($h, y$)
    	\end{algorithmic}
        \end{algorithm}
    
\section{BOSS Enhancements}
\label{enhancements}   
    Our changes to BOSS mainly focus on the ensemble technique of the classifier, which is computationally expensive and unpredictable. Ensembling has been shown to be an essential component of BOSS, resulting in significantly higher accuracy~\cite{large19dictionary}.  We assess whether we can replace the current ensemble mechanism with a more stable and efficient scheme without a significant reduction in accuracy.  
    
    The ensemble created through its parameter search is unlike more standard ensemble based classifiers such as Random Forest~\cite{breiman01randomforest}, in which the number of classifiers in the ensemble is set beforehand. We investigate a more traditional ensemble method of fixing the number of base BOSS classifiers, then injecting diversity through randomising the parameter settings. To differentiate the two ensemble methods we will refer to BOSS using randomised parameters as RBOSS. This introduces a new parameter, $k$, the fixed ensemble size.
    
    Randomising the parameter settings may not inject enough diversity. Hence, we also test a version of RBOSS that creates diversity by using a subsample of the train data for each classifier. This has the additional benefit of reducing the train size for each ensemble member, which results in a reduction to the overall time and space complexity of the classifier and speeds up any train accuracy estimates.

We further assess an alternative voting scheme based on weighted probablities rather than votes, which is currently employed in alternative HIVE-COTE components and is referred to as the Cross-validation Accuracy Weighted Probabilistic Ensemble (CAWPE, previously known as HESCA)~\cite{large17hesca}.
    CAWPE is an accuracy weighted voting scheme based on the probability distribution from a cross validation on the train data.
    Additionally a weight $\alpha$ is applied to the weightings as an exponent, intending to to attenuate differences in accuracy.
    
    A further problem with BOSS is that it is very memory intensive, since each classifier must save the transformed data for the nearest neighbour classifier.    To mitigate this we look into replacing the nearest neighbour classifier with an alternative which would not require the storage of transformed data while retaining accuracy. Given the similarity in structure of BOSS to tree based ensembles, it is natural to assess decision tree classifiers. We assess C4.5 and the random tree used by random forest.       We also try  using a support vector machine and logistic regression (also used in WEASEL).
    
    The fixed size of RBOSS makes it possible to replace the ensemble size parameters $k$ with an amount of time.
    By doing this we can contract the classifier to build as many individual BOSS classifiers in the given amount of time, thus giving us predictable amount of build time regardless of dataset characteristics.
    However, for small problems this could easily result in a very large number of base classifiers with a huge memory requirement.     A simple way to counteract this is to add a maximum number of classifiers that can be built. To accommodate long build times for large problems, we have introduced check-pointing to RBOSS.
    When enabled, check-pointing will save the current state of the classifier periodically to file, allowing the classifier to start building from that point in the event building ceases.

    While discarding the accuracy estimate would save the substantial amount of time spent doing cross-validation, it was quickly found that many datasets rely on this to select specific parameters they perform well on.
    As such we look into speed ups while retaining this estimate of performance for some of our classifier variants.
    We retain the incremental building of random parameter sets from RBOSS while doing this, but set a max limit to the number of classifiers that can be in the ensemble $s$.
    If the ensemble is full the current lowest accuracy classifier is replaced by any newly built classifier with higher accuracy.
    This has the added benefit of reducing the amount of classifiers each with transformed data that have to be stored.
    Improvements to the accuracy estimate could also be made, as the goal of the estimate is to weed out poor performing classifiers it does not have to be precise.
    We investigate the effects of limiting the current leave-one-out cross-validation, taking a subsample of train instances to create the accuracy estimate rather than the full set. 

\subsection{Best Performing RBOSS Variation}

    The best performing and definitive version of RBOSS found through our experiments is a combination of the above enhancements. 
    The classifier retains an accuracy estimate to filter bad classifiers while including CAWPE weights and train data subsampling.
    RBOSS builds base BOSS classifiers until the ensemble size $k$ is reached, using a unique parameter set for each.
    Leave-one-out cross-validation is used to generate probabilities for the CAWPE weightings and to estimate accuracy.
    As both CAWPE and the accuracy estimate require this there is minimal overhead for the addition of one when the other is present.
    If the current number of classifiers is over the max ensemble size $s$, the current classifier with the lowest accuracy is replaced if any new classifier has an accuracy value over it.
    The option to replace the ensemble size with a time limit through contracting is also available, continually building new classifiers until it runs out of time or no parameter sets remain.
    The process for building RBOSS is described in algorithm \ref{cawperboss}.

    \begin{algorithm}[]
        	\caption{buildWeightedRBOSS(A list of $n$ cases length $m$, ${\bf T}=({\bf X,y})$)}
        	\label{cawperboss}
        	\begin{algorithmic}[1]
        \REQUIRE the ensemble size $k$, the max ensemble size $s$
        		\STATE Let ${\bf C}$ be a list of $s$ BOSS classifiers $({\bf c}_1,\ldots,{\bf c}_s)$
        		\STATE Let ${\bf E}$ be a list of $s$ classifier weights $({\bf e}_1,\ldots,{\bf e}_s)$
        		\STATE Let ${\bf R}$ be a set of possible BOSS parameter combinations
        		\STATE $i \leftarrow 0$
        		\STATE $lowestAcc \leftarrow \infty, lowestAccIdx \leftarrow \infty$
        		\WHILE {$i < k$ AND $|{\bf R}| > 0$}
        		    \STATE $r \leftarrow$ rand$(1,|{\bf R}|)$
        		    \STATE $[l,a,w,p] \leftarrow R_r$
        		    \STATE ${\bf T'} \leftarrow$ subsampleData(${\bf T}$)
        		    \STATE $cls \leftarrow$ buildBaseBOSS(${\bf T'},l,a,w,p$)
        		    \STATE ${\bf P} \leftarrow$ LOOCV($cls$) \COMMENT{ {\em train data probabilities}}
        		    \STATE $acc \leftarrow$ accuracy(${\bf P}$)
        		    \IF{$i < s$}
        		        \IF{$acc < lowestAcc$}
        		            \STATE $lowestAcc \leftarrow acc$, $lowestAccIdx \leftarrow i$
        		        \ENDIF
        		        \STATE $c_i \leftarrow cls$, $e_i \leftarrow$ CAWPEWeight(${\bf P}$)
        		    \ELSIF{$acc > lowestAcc$}
        		        \STATE $c_{lowestAccIdx} \leftarrow cls$, $e_{lowestAccIdx} \leftarrow$ CAWPEWeight($cls$)
        		        \STATE $[lowestAcc,lowestAccIdx] \leftarrow$ findNewLowestAcc(${\bf C}$)
        		    \ENDIF
        		    \STATE $i \leftarrow i+1$
        		\ENDWHILE
        	\end{algorithmic}
        \end{algorithm}

\section{Results}
\label{results}
    For this comparison the well known dictionary based classifiers bag of patterns (BOP)~\cite{lin12bagofpatterns}, BOSS~\cite{schafer15boss} and WEASEL~\cite{schafer2017fast} are included as well as our different variations of RBOSS.
    As a convention for graphs and tables the size of the ensemble will be noted next to the classifier name along with its difference from the base RBOSS, with RBOSS100 denoting an ensemble size of 100.
    Each of the classifiers are tested on 30 random resamples from the 114 datasets without missing values on the UCR repository unless noted otherwise, with each resample being the same train test split for all classifiers tested.
    
    Comparison of the performance of classifiers is done using pairwise Wilcoxon signed rank tests and cliques are formed using the Holm correction, following recommendations from \cite{benavoli16pairwise} and \cite{garcia08pairwise}.
    Classifiers in the same clique show no significant difference from each other in the results from the tested datasets.
    The resulting ranks and cliques are represented in a critical difference diagram, with cliques shown as a solid bar.
    
    Most of the replacement classifiers for the nearest neighbour failed to complete the full 114 datasets, with the exception of the random tree.
    On the problems that completed successfully they did not have the desired effect, most failing to reducing the space requirement sufficiently and drastically increasing build time.
    In addition to this the accuracy of the classifiers was poor, with some having over 10\% lower accuracy on average than nearest neighbour with the same ensemble size and classifier parameter sets. 
    
    The pool of parameters we randomly pick from for RBOSS is the same as BOSS searches, shown in table~\ref{BOSSparameters}.
    While dependant on the size of the series, the window length parameters has the largest range of values by a large margin.
    By default BOSS sets the maximum possible window length value to be the same as the series length, essentially creating a histogram with 1 word if the value is chosen.
    In most cases this would produce a bad individual classifier, but while BOSS can filter these out, RBOSS cannot.
    Comparing changes for this we found setting the maximum length to half was close to 1\% more accurate on average for RBOSS.
    As such we replace the original maximum window length with half the length of the series $m/2$ rather than the full length $m$ for classifiers that do not create an accuracy estimate.
    
    For our different RBOSS variants we use a default ensemble size of 100 to test the variation in accuracy from different enhancements to the base classifier.
    We also include a base RBOSS classifier of ensemble size 250 to show the effect of increasing the ensemble size on performance.
    On our subsample experiment we take a random 70\% of the full dataset to build each individual classifier.
    For our contract RBOSS we test two time limits, 10 minutes and 1 hour.
    We include a limit of 500 classifiers that can be built during the contract.
    This limit is hit by many datasets and is required to prevent a premature exit due to the required amount of memory.
    
        \begin{figure}[!ht]
        	\centering       
            \includegraphics[scale=0.4, width=12.5cm]{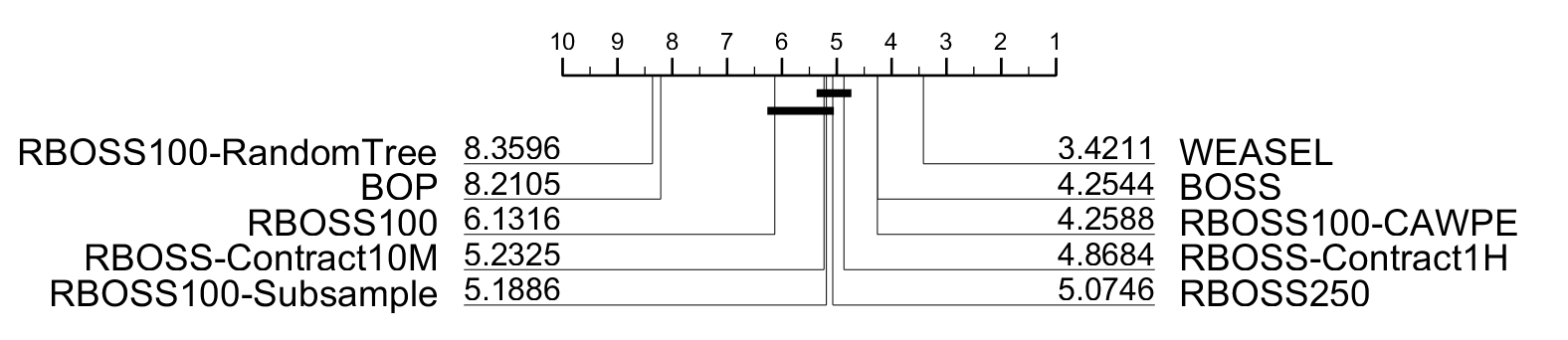}
               \caption{Critical difference diagram showing accuracy ranks and cliques for the dictionary based classifiers and RBOSS variants.}
               \label{critdiffacc}
        \end{figure}
        
    The critical difference diagram for our comparison of RBOSS variants and current dictionary based classifiers is shown in figure~\ref{critdiffacc}.
    As expected WEASEL and BOSS are the top performers when it comes to accuracy rank, with WEASEL top ranked and significantly better than the other classifiers, confirming published results.
    While our weighted RBOSS classifier is very close to BOSS in rank, all of the tested RBOSS variants perform significantly worse than BOSS.
    
    However out experiments showed a wide range in ranking from our different varieties of RBOSS, from which we could choose the better performing improvements to possibly close the performance gap. 
    One of the features of RBOSS is more control over the size of the ensemble through setting it directly and contracting, and as would be expected this value has a large effect on performance.
    With a base RBOSS classifier an ensemble size of 100 or below results in a significantly worse classifier than one of size 250 or contracted for 1 hour.
    This consistent increase in preference using a longer contract time/ensemble size shows a more controllable trade off in train time and space for accuracy for RBOSS.
    The CAWPE variant of RBOSS was the best performer out of the ones we tested, performing significantly better than other RBOSS classifiers.
    Having weighting for the classifiers had a larger effect than ensemble size, performing better than the default RBOSS classifiers with higher size.
    While extra time is taken the retrieve the weights, valuing the ensembles classifiers on them proves crucial to retaining accuracy.
    Taking a subsample of the data also improved the performance of the classifier, indicating there may be some issues in the base version with the diversity of the classifiers.
    
    Looking at the individual dataset performance there are several for which BOSS performs much better than RBOSS, most apparent in SyntheticControl where its accuracy is 0.61 higher and CBF where it is 0.646 higher.
    Upon looking closer at the CBF dataset to understand why this might be, we found that of the 49 ensemble members BOSS had picked for its ensemble all but 6 have a word length of 8. 
    The random selection in the RBOSS ensemble creates more variety in its parameters, and when some parameter values lead to a very bad classifier, RBOSS suffers greatly in performance. 
    A similar pattern appears in other datasets where RBOSS performed poorly.
    While this is slightly mitigated by weighting, the mass of poor performing classifiers on the dataset still brings down accuracy compared to BOSS.
    This shows one of the strengths of BOSS in its ability to perform well on datasets where specific parameters are required for good performance.
    
        \begin{figure}[!ht]
        	\centering       
            \includegraphics[scale=0.4, width=12.5cm]{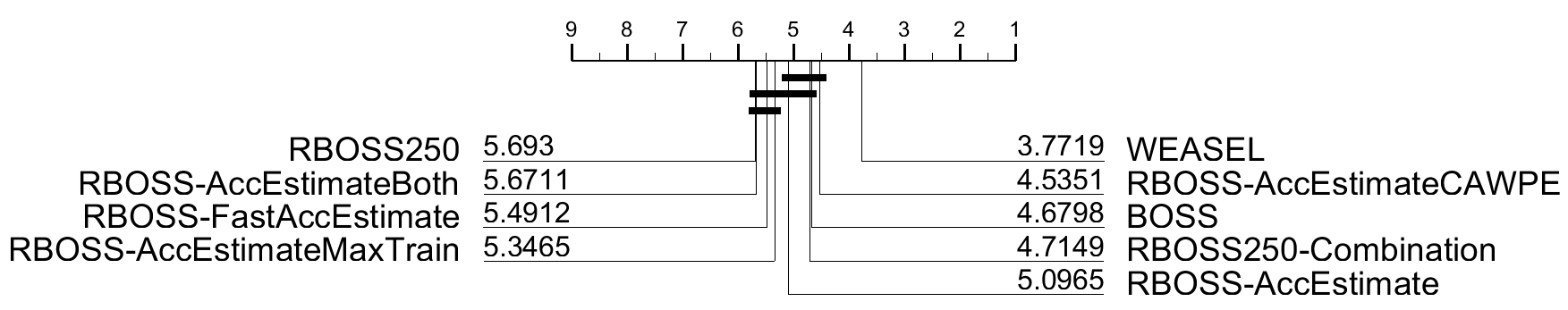}
               \caption{Critical difference diagram showing accuracy ranks and cliques for the dictionary based classifiers and second round of RBOSS variants.}
               \label{critdiffacc2}
        \end{figure}
    
    This major drop in accuracy due to bad ensemble members led us to run a second round of experiments including a max ensemble size and accuracy estimate to act as a filter.
    We test multiple versions of a filtered RBOSS, each with a max ensemble size of 50 and set to build 250 classifiers using unique parameter sets.
    We look into a larger reduction of train size for each classifier in an aim to further reduce space complexity by taking a stratified subsample of size of maximum 500. This provides a hard constraint for the memory requirements. 
    In our test to limit the leave-one-out cross-validation we select random 50 instances for each class value in a dataset from which to estimate performance.
    We include a classifier using both this limited estimate and max train size to gauge how much performance is lost using these techniques.
    
    While individually significantly different from BOSS a number of experiments from the first round improved performance on our base case.
    We test two classifiers using the results found.
    The first is a combination of CAWPE weights and a 70\% train subsample using an ensemble size of 250.
    The second uses CAWPE and subsampling alongside our base filtered RBOSS.
    We additionally modified CAWPE to use the leave-one-out estimates produced instead of preforming its own cross-validation, providing a slight speed up to the process.
    
    \begin{figure}[!ht]
        	\centering       
            \includegraphics[scale=0.45]{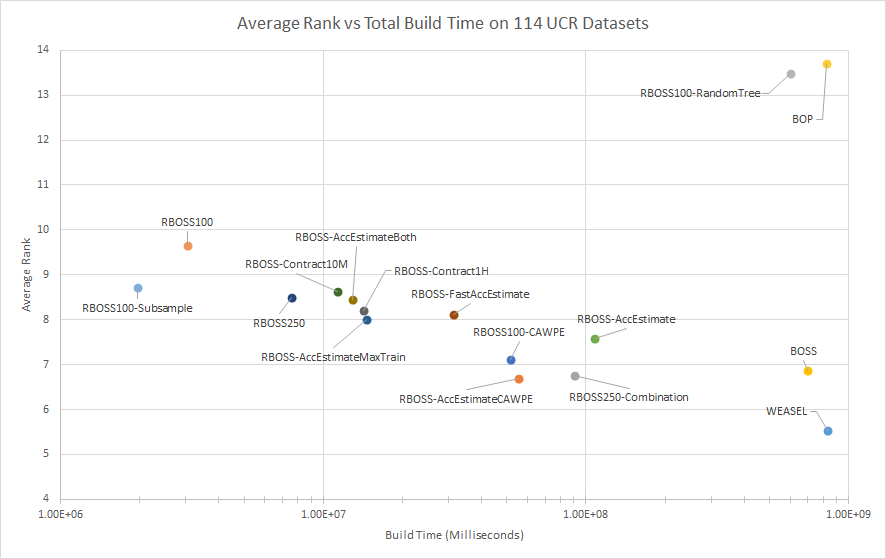} 
               \caption{Average build time on a log scale against average rank for each classifier, with a lower value for both axis being a better performance.}
               \label{resultstraintime}
        \end{figure}
    
    Figure~\ref{critdiffacc2} shows a critical difference diagram of the results from our second round of experiments.
    An immediate improvement can be seen with the accuracy estimate filter as all of the classifiers which make use of it are not significantly different than BOSS.
    This is especially seen in our two CAWPE classifiers, which rank close to BOSS with the accuracy estimate filter version ranking higher.
        
    The enhancements in RBOSS have shown to not significantly reduce the performance of the classifier compared to BOSS when retaining an accuracy estimate. 
    Figure~\ref{resultstraintime} shows the total train time on the 114 datasets for each of the classifiers against its ranking comparing all classifiers tested.
    The results for the build time experiment showed that on average RBOSS of size 250 builds around 87 times faster than standard BOSS, with RBOSS totaling 133 minutes to build and BOSS 11620 minutes.
    The cross-validation required to find the weights for our filtered CAWPE RBOSS adds a significant amount of time to build compared to default RBOSS.
    However it is still over 12 time faster than BOSS at 926 minutes, while retaining the performance that the unfiltered version does not.
    
    \begin{table}[h]
            \centering
			\caption{Average large dataset performance by classifier using Accuracy, Area Under the Receiver Operating Characteristic (AUROC) and Negative Log Likelihood (NLL). Included is total build time over all datasets relative to BOSS.}
			\begin{tabular}{r c c c c c}
				Classifier & AccRank & AUROC & NLL & Build Time \\
				\hline
                BOSS & 4.6875 & 0.953 & 0.7923 & 100\% \\ 
                WEASEL & 2.1875 & 0.9529 & 0.9753 & 85.48\% \\ 
                RBOSS250 & 6.375 & 0.933 & 1.0432 & 0.79\% \\ 
                RBOSS250-Combination & 4.9375 & 0.9446 & 0.9455 & 12.77\% \\
                RBOSS-AccEstimate & 4.9375 & 0.9523 & 0.8023 & 15.01\% \\ 
                RBOSS-AccEstimateMaxTrain & 6.34375 & 0.9517 & 0.8791 & 1.29\% \\ 
                RBOSS-FastAccEstimate & 5.3125 & 0.9521 & 0.8104 & 3.65\% \\ 
                RBOSS-AccEstimateBoth & 6.15625 & 0.9515 & 0.8846 & 1.07\% \\
                RBOSS-AccEstimateCAWPE & 4.0625 & 0.9558 & 0.7951 & 7.7\% 
			\end{tabular}
			\label{resultstable}
		\end{table}
    
    Our results so far show promise, but due to the volume of small problems in the archive the effect of our changes on problems that it matters for the most may not be shown clearly.
    As such we take a closer look at the results for 16 problems on which BOSS takes over an hour to build.
    Table~\ref{resultstable} shows the results of our second round classifiers on these datasets.
    On the full dataset setting a max train size performed better than our fast accuracy estimate, the opposite is true when filtering large datasets only.
    Compared to the default filtered RBOSS a significant amount of time is saved by both speed up attempts, though some performance is lost.
    The filtered CAWPE version of RBOSS remains the definitively best performing version of the classifier, ranking higher than BOSS with a large speed up in build time.
    
\begin{figure}[!ht]
        	\centering
            \includegraphics[scale=0.5]{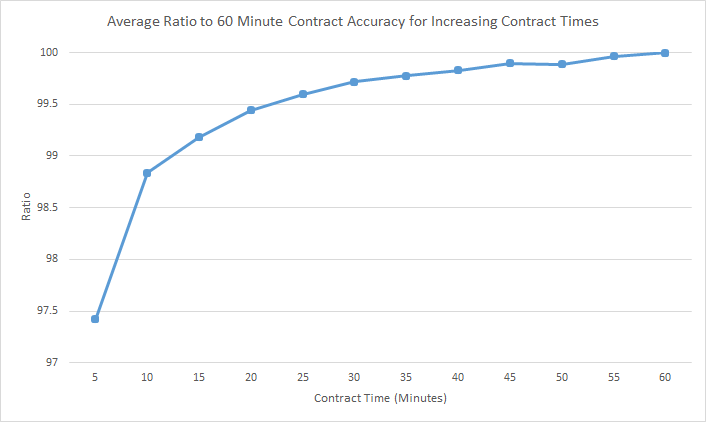}
              \caption{Average accuracy value for a range of contract times on 16 large datasets. Times                 range from 5 to 60 minutes linearly spaced in increments of 5.}
              \label{largedatacontract}
        \end{figure}

    To demonstrate the effectiveness of building RBOSS using a time contract figure~\ref{largedatacontract} shows the change in accuracy over a range of $t$ values on the 16 large problems.
	As shown an increase in time the classifier contracted for increases accuracy on average, though this increase slows as the parameter search progresses. The versatility of being able to build the best classifier within a train time limit is a large usability boost to the classifier, making this aspect much more predictable.
		
    \subsection{Whale Acoustics Use Case}
    
    Recently, the protection of endangered whales has been a prominent global issue. Being able to accurately detect marine mammals is important to monitor populations and provide appropriate safeguarding for their conservation. North Atlantic right whales are one of the most endangered marine mammals with as few as 350 individual remaining in the wild~\cite{kraus2005north}. 
    We use a dataset from the Marinexplore and Cornell University Whale Detection Challenge\footnote{https://www.kaggle.com/c/whale-detection-challenge/data} that features a set of right whale up-calls. Up-calls are the most commonly documented right whale vocalisation with an acoustic signature of approximately 60Hz-250Hz, typically lasting 1 second. Right whale calls can often be difficult to hear as the low frequency band can become congested with anthropogenic sounds such as ship noise, drilling, piling, or naval operations~\cite{cox2006understanding}. 
    Each series is labelled as either containing a right whale or not with the aim to correctly identify the series that contain up-calls. 
        
    Previous work has been done in classifying the presence of whale species using acoustic data with whale vocalisations~\cite{dugan2010north,shamir2014classification}. However, TSC approaches have not been applied.
   While the type of data is important to whether an algorithm will perform well on it, the most important characteristic of this dataset to us is its size.
    The dataset contains 10,934 train cases and is sampled at 2kHz for a series length of 4000.     This exceeds the largest train set from ElectricDevices of 8926 and series length from HouseTwenty of 3000 in the 128 UCR datasets~\cite{dau18archive} (we will donate this data to the archive for the next release). A combination of these factors causes a problem for many standard approaches which would find it difficult to build a model on this data with reasonable time and memory.
   
    We test the classifiers that make up the HIVE-COTE ensemble~\cite{lines18hive} on this dataset as well as RBOSS.
    Only three of the five classifiers that make up HIVE-COTE were able to process the dataset in a reasonable amount of time, the failures being BOSS and the Elastic Ensemble (EE)~\cite{lines15elastic} which went over our processing clusters 28 day limit for its large memory queue.
    Along with our variants of RBOSS the other algorithms that successfully built are the Random Interval Spectral Ensemble (RISE)~\cite{lines18hive}, Shapelet Transform (ST)~\cite{lines12shapelet} (with a contracted time limit of five days~\cite{bostrom17binary}) and Time Series Forest (TSF)~\cite{deng13forest}.
        
        \begin{table}[ht]
            \centering
			\caption{Accuracy and build time in hours for each of the HIVE-COTE components and different RBOSS configurations.}
			\begin{tabular}{r c c}
				Classifier & Accuracy & Build Time (Hours) \\
				\hline
				RISE & 0.7859 & 141.96  \\
				ST & 0.818 & 20.05  \\
				TSF & 0.7712 & 16.23  \\
				RBOSS50 & 0.8084 & 1.17  \\
				RBOSS100 & 0.8175 & 1.83 \\
				RBOSS-Contract10M & 0.8017 & 0.18 \\
				RBOSS-Contract1H & 0.8104 & 1.01  \\
				RBOSS-Contract4H & 0.8145 & 2.93  \\
				RBOSS-AccEstimate & 0.8129 & 227.94 \\
				RBOSS-AccEstimateBoth & 0.7823 & 0.29 \\
				RBOSS100-Combination & 0.8053 & 70.64 \\
				RBOSS-AccEstimateCAWPE & 0.8114 & 119.39
			\end{tabular}
			\label{whalestable}
		\end{table}
		
	Table~\ref{whalestable} shows the accuracy and build time on the whales dataset each of the COTE classifiers and some variants of RBOSS.
	While BOSS was unable to build the dataset in a reasonable amount of time, all RBOSS variants succeeded, with some being significantly faster than any of the COTE classifiers.
	The best performing of the classifiers is ST, with ensemble size 100 RBOSS coming close with over 10 times less build time.
	With a large dataset the amount of memory required is uncertain with the varying parameters used.
	Check-pointing shows its usefulness in these situations, allowing for quick recovery in situations where not enough is allocated.
	While EE fails to build, with the alternative BOSS the dictionary portion no longer becomes an issue in running HIVE-COTE for the dataset.
	
	While successful compared to BOSS in that we were able to get results, some of the remaining issues in each of our experimental variants are amplified in the data.
	The 4 hour contract ended early with 155 total classifiers built, as it no longer became feasible to let it continue, with it requiring over 350GB of RAM at the stopping point.
	While this is better for the classifiers with a max ensemble size of 50, a large requirement is still requires with around 70-110GB depending on if subsampling is used.
	The variants which require an accuracy estimate were dramatically slower than ones without compared to the difference on the average data, displaying the remaining scaling issue that comes with cross-validating.
	Alongside this with classifiers that subsampled the data performance was poor compared to the simpler RBOSS versions, showing that these enhancements do not boost performance for all datasets.

\section{Conclusion}
\label{conc}

    We present a more scalable version of the BOSS classifier, replacing its current ensemble.
    The replacement of the parameter search with randomly selected parameter sets provides a considerable speed up, and the introduction of subsampling for increased diversity and CAWPE weights leads to increased performance on average.
    While we tested different versions of the classifier, we think of this setup as the definitive version of RBOSS and will be the version used in future works when referring to RBOSS.
    While removing an accuracy estimate increases this speed up significantly it comes at the cost of a significant amount of accuracy.
    Our inclusion of a fixed ensemble size, the ability to contract the build time and save progress with check-pointing make the classifier more robust and more predictable.
    
    While WEASEL performed significantly better than all classifiers tested it also comes in as the slowest classifier to build on average, performing slightly better than BOSS on larger datasets however.
    While this points to it being a suitable BOSS replacement on smaller datasets, it runs into the same problems as BOSS as the dataset size rises.
    Alongside WEASEL we recommend the use of our filtered CAWPE-RBOSS as a possible replacement for BOSS in situations where WEASEL fails to build in a reasonable amount of time.
    
    While our better performing changes to RBOSS significantly improve scalability compared to BOSS, datasets which present possibly several hundred thousand instances may still cause an issue in requirements for space and time.
    Potential solutions for this may be found improved applications of our fast accuracy estimate and max train size shown in out second round of experiments.
    If space usage for a dataset given individual classifier parameters could be estimated, a contract for memory could be introduced, setting a max train size to fit around this limit.
    Investigations into the feasibility of this and any affect on the classifiers performance could be an interesting future work in further improving dictionary based scalability.
    
\section*{Acknowledgements}{
 This work is supported by the UK Engineering and Physical Sciences Research Council (EPSRC) iCASE award T206188 sponsored by British Telecom. The experiments were carried out on the High Performance Computing Cluster supported by the Research and Specialist Computing Support service at the University of East Anglia.
}

\bibliographystyle{plain}
\bibliography{main,TSCMaster} 

\end{document}